\documentclass[letterpaper, 10 pt, conference]{ieeeconf}  
\IEEEoverridecommandlockouts

\let\labelindent\relax
\usepackage{cite}
\usepackage{amsmath}
\usepackage{amssymb}
\DeclareMathOperator*{\argmax}{arg\,max}

\usepackage{wrapfig}
\usepackage{booktabs}
\usepackage{multicol}
\usepackage{multirow}
\usepackage{tabularx}
\usepackage{diagbox}
\usepackage{graphicx}
\usepackage{subcaption}
\usepackage{array}
\usepackage{cleveref}
\usepackage{xcolor}
\usepackage{tikz}
\usepackage{verbatim}
\usepackage{listings}
\definecolor{mygreen}{RGB}{28,172,0} 
\definecolor{mylilas}{RGB}{170,55,241}

\usepackage{matlab-prettifier}
\usepackage{color}
\usepackage[shortlabels]{enumitem}

\usepackage{algorithm}
\usepackage[noend]{algpseudocode}
\makeatletter
\def\BState{\State\hskip-\ALG@thistlm}
\makeatother
\newcommand{\E}{\mathbb{E}}

\author{Haozhe Lei$^{1}$ and Quanyan Zhu$^{1}$ 
\thanks{$^{1}$The authors are with the Department of Electrical and Computer Engineering, New York University, Brooklyn, NY, 11201, USA.
        {\tt\small hl4155@nyu.edu, qz494@nyu.edu}.}%
}
\begin{document}

\title{\LARGE\bf Neurosymbolic Meta-Reinforcement Lookahead Learning Achieves Safe Self-Driving in Non-Stationary Environments
}

\maketitle

\begin{abstract}
    In the area of learning-driven artificial intelligence advancement, the integration of machine learning (ML) into self-driving (SD) technology stands as an impressive engineering feat. Yet, in real-world applications outside the confines of controlled laboratory scenarios, the deployment of self-driving technology assumes a life-critical role, necessitating heightened attention from researchers towards both safety and efficiency. To illustrate, when a self-driving model encounters an unfamiliar environment in real-time execution, the focus must not solely revolve around enhancing its anticipated performance; equal consideration must be given to ensuring its execution or real-time adaptation maintains a requisite level of safety. This study introduces an algorithm for online meta-reinforcement learning, employing lookahead symbolic constraints based on \emph{Neurosymbolic Meta-Reinforcement Lookahead Learning} (NUMERLA). NUMERLA proposes a lookahead updating mechanism that harmonizes the efficiency of online adaptations with the overarching goal of ensuring long-term safety. Experimental results demonstrate NUMERLA confers the self-driving agent with the capacity for real-time adaptability, leading to safe and self-adaptive driving under non-stationary urban human-vehicle interaction scenarios.
\end{abstract}


\section{Introduction}
The application of machine learning (ML) in self-driving (SD) technology represents a marvel of engineering, enabling vehicles to process an array of sensor inputs in real-time, interpret complex surroundings, and execute actions with a precision that was once relegated to the realm of science fiction. Recent advances in the field of machine learning, as evidenced by works such as \cite{ross11dagger,mnih2015DQN,tao_confluence}, have triggered a significant surge of curiosity and investigation into the realm of learning-driven SD \cite{kiran21drl_ad}. This application has arisen in vehicles that can correctly work through known cityscapes, anticipate pedestrian behavior, and interact perfectly with other vehicles, all while following traffic rules and optimizing fuel efficiency. Nevertheless, beyond controlled experimental setups, the inherent unpredictability of artificial intelligence (AI) becomes evident when a self-driving vehicle confronts a new and unfamiliar situation. In such instances, the system's performance might deteriorate or lead to a crash when encountering unanticipated scenarios on real-world roads. The inaugural instance of a pedestrian fatality attributed to autonomous vehicles surfaced in 2018, when a self-driving Uber vehicle collided with a pedestrian crossing an intersection in Tempe, Arizona, during the nighttime \cite{wakabayashi2018self}. This tragic event highlights the critical importance of improving the safety and adaptability of autonomous driving systems. Considering such challenges, a pertinent question arises: How can advances in technologies and methodologies help enhance the capability of autonomous vehicles to operate safely and reliably in diverse and complex environments?

\begin{figure}[h]
    \centering
   \includegraphics[width=0.49\textwidth]{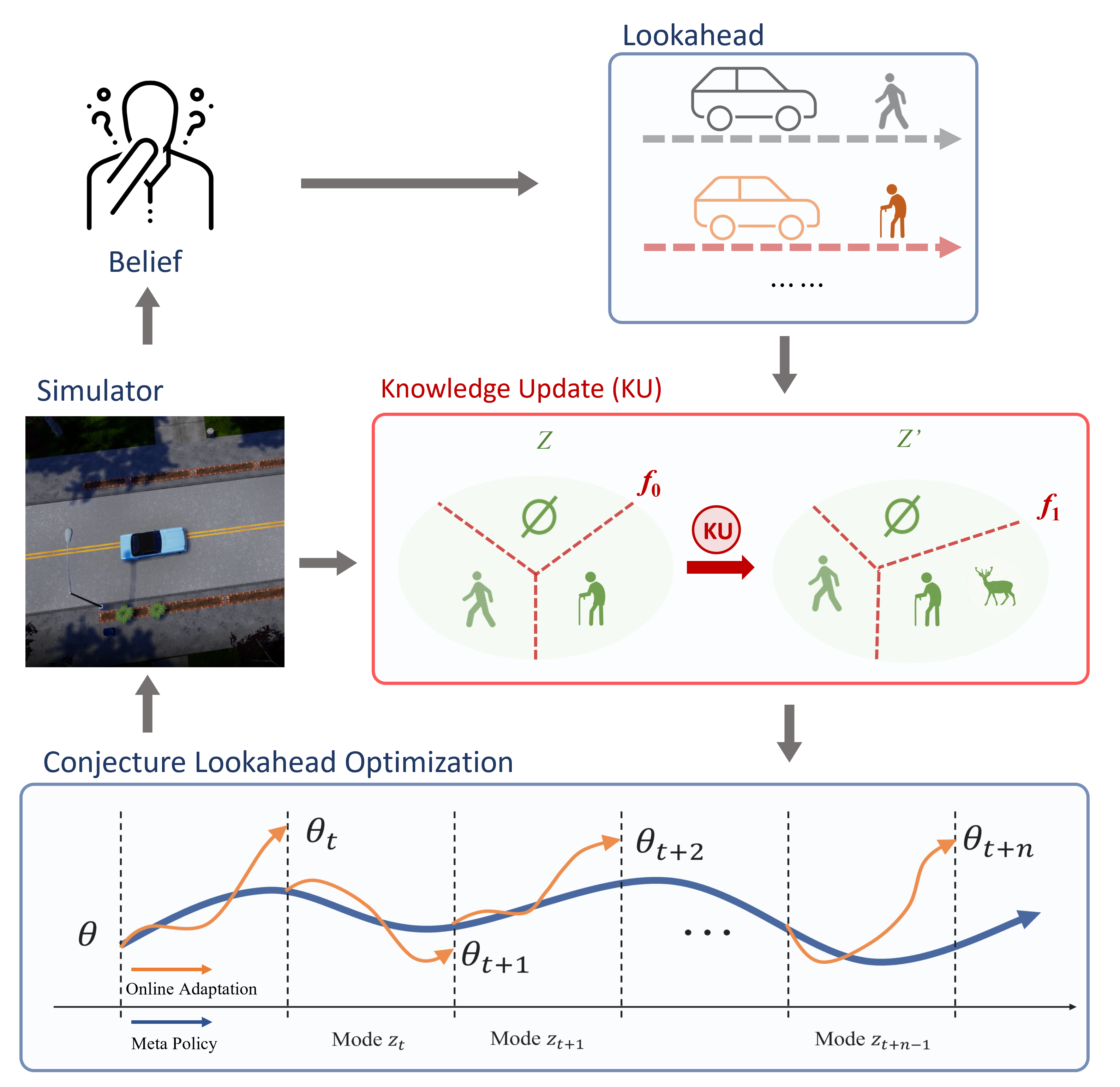}
    \caption{An illustration of Neurosymbolic Meta-Reinforcement Lookahead Learning. When driving in a changing environment, the agent first uses observation from the environment to calibrate its belief at every time step about the mode. Based on its belief, the agent conjectures its performance in the future within a lookahead horizon. Then, using this conjecture, the agent searches in its knowledge to find suitable safety constraints. In the meantime, the knowledge of the agent will update itself by symbolic safety constraint adaptation if needed. The policy is adapted through conjectural lookahead optimization with safety constraints, leading to a suboptimal (empirically) online control with a long-term safety guarantee.}·
    \label{fig:system_diagram}
\end{figure}

In reinforcement learning (RL), stochastic policies are commonly employed in partially observable environments, such as robotics or autonomous driving (no matter in Q-learning methods \cite{mnih2015DQN, tao_multiRL} or policy-based methods \cite{sutton_PG,Tao_blackwell}), where agents encounter sensor noise and incomplete information. During offline training, stochastic policies offer various advantages, including increased robustness against uncertainty and environmental variations, improved exploration capabilities, and compatibility with policy search algorithms like evolutionary strategies or Monte Carlo (MC) methods \cite{silver10mcpomdp}. However, the limited generalization ability prevents RL from wide application in real SD systems when encountering nonstationary environments different from their training time. This drawback also makes the stochastic policies even more unstable in the life-critical execution.

Enhancing the adaptability of reinforcement learning (RL) policies comprises the objective of meta-reinforcement learning (meta-RL), which attempts to discover a meta-policy capable of adeptly adjusting and delivering satisfactory performance across a spectrum of environments \cite{finn18adapt_dy}. While prior works \cite{finn18adapt_dy, finn2017model, rakelly19a, finn20_task_agnostic} have dedicated significant efforts to this pursuit, many of them continue to rely on offline methodologies. These approaches display the capacity to adapt to a diverse array of tasks within environments they are exposed to during their training in offline settings. The practical implementation of online machine learning often faces challenges due to its time-sensitive nature. Processing and updating models in real-time can be demanding, making time a critical factor when deploying such systems in real-world applications. Beyond the time constraints, another significant challenge arises: the assurance of policy processing safety remains an ongoing concern in these methods.

In conjunction with the real-time adaptation capability, several researchers have incorporated safety-centric learning to support policy robustness. For instance, in \cite{anderson2020neurosymbolic}, an approach leveraging symbolism is proposed to formulate distinct safety policies tailored to various state partitions. Similarly, \cite{alshiekh2017safe} presents a method that constructs a shield for actions based on observation inputs, ensuring the safety of each individual step. Notably, neither of these approaches accounts for the dynamic nature of the environment. This implies that in scenarios where the initial environmental observations are incomplete or where the environment is subject to change, the effectiveness of the safety mechanisms might diminish.

 \textbf{Our Contributions} In response to the dual challenges of limited real-time adaptation capabilities and the quest for safety assurance, this study introduces an algorithm for online meta-reinforcement learning, employing lookahead symbolic constraints based on \emph{Neurosymbolic Meta-Reinforcement Lookahead Learning} (NUMERLA). The underlying principle of NUMERLA is to facilitate secure real-time learning by continually updating safety constraints. The core idea involves employing logical statements as safety constraints for the process of secure online meta-adaptation learning (OMAL) [See \Cref{sec:obj_NUMERLA}]. These constraints are iteratively refined in a forward-looking manner during the online execution [See \Cref{eq:NUMERLA}]. This lookahead updating mechanism balances the efficiency of online adaptations with the overarching goal of ensuring long-term safety.

 In summary, the main contributions of this work include: 1) conceptualizing the challenge of acquiring adaptive strategies in a dynamic environment characterized by symbolic safety constraints; 2) introducing an ensure-safe Real-Time OMAL algorithm, which builds upon the principles of Neurosymbolic Meta-Reinforcement Lookahead Learning (NUMERLA); 3) experimental results demonstrating that NUMERLA enables the self-driving agent with the capacity for real-time adaptability, leading to safe and self-adaptive driving under non-stationary urban human-vehicle interaction scenarios.

\section{Definition and Model Structure of NUMERLA}
\subsection{Meta Reinforcement learning}
RL is a field that focuses on solving problems within a stationary environment called a Markov Decision Process (MDP). Considering $z_t\in \mathcal{Z}$ is the environment mode or latent variable hidden from the agent at time $t$. Let $s_t\in\mathcal{S}$ and $a_t\in\mathcal{A}$ be the state input and the control action at time $t$. In the context of RL, we often encounter situations where the underlying conditions remain stable throughout a decision-making period, known as $H$. This means that the parameters that define the environment remain unchanged as time progresses (i.e., $z_t=z$). This characteristic is known as "stationarity," which allows us to consider a specific class of policies known as Markov policies \cite{puterman_mdp}. These policies, denoted as $\pi: \mathcal{S} \rightarrow \Delta(A)$, depend only on the current state.

Denote a neural network-based policy by $\pi(s,a;\theta)$, where $\theta \in \Theta \subset \mathbb{R}^d$ ($d$ represents the dimension of parameters in the neural network). This choice aligns well with scenarios where state information is collected from sensors. And the evaluation of the choices will be given by reward function $r_t=r(s_t,a_t)$. The aim of RL is to tackle a problem in the realm of stationary MDPs, where the environment determines the best policy that maximizes the expectation of cumulative rewards in a fixed environment $z$. These rewards are accounted for over time using a discount factor $\gamma$ ($0 < \gamma \leq 1$):
\begin{align}
    \max_{\theta} J_z(\theta) := \mathbb{E}_{P(s_{t+1} | s_t,a_t; z), \pi(s_t,a_t; \theta)} \left[ \sum_{t=1}^H \gamma^t r(s_t, a_t) \right].\tag{RL}
    \label{eq:rl}
\end{align}
Here, the transition $P(s_{t+1}|s_t,a_t;z_t)$ tells how likely the agent is to observe a certain state $s_{t+1}$ with control action $a_t$ under the current mode conditions $z_t$. Since we are using the fixed $z_t=z$, it can be reduced to $P(s_{t+1}|s_t,a_t;z)$.

Traditional meta-RL is explored in works like \cite{finn18adapt_dy,rakelly19a,finn20_task_agnostic}. These methods aim to discover a meta policy denoted as $\theta$, along with an adaptation mapping called $\Phi$. The objective is to attain favorable rewards across various environments by updating the meta policy $\theta$ through $\Phi$ within each environment. In simpler terms, they try to train a good meta policy using offline methods. Instead of treating meta-learning as a fixed optimization problem, we propose learning the meta-adaptation process in real-time. This implies that the agent adjusts its adaptation strategies continuously based on its observations. Essentially, our approach enables the agent to adapt to changing environments. The following paragraph formally defines the problem of online meta-adaptation learning (OMAL).

Let $\mathcal{I}_t=\{s_t,a_{t-1},r_{t-1}\}$ be the set of agent's observations at time $t$, referred to as the information structure \cite{tao_info}. The online adaptation mapping relies on the online observations $\cup_{k=1}^t\mathcal{I}_k$. Then the meta adaptation mapping at time $t$ is defined as $\Phi_t(\theta):=\Phi(\theta,\cup_{k=1}^t\mathcal{I}_k)$. The adaptation mapping $\Phi$ adapts the meta policy $\theta$ to a new policy fine-tuned for the specific $z$ at time step $t$ based on the agent's observations $\mathcal{I}_t$. Under this circumstance, we will use expected reward $r_t^\pi(s_t;\theta):=\E_{a\sim \pi(\cdot|s_t;\theta)}[r_t(s_t,a)]$ as our new objective in the function shown below:
\begin{align}
    &\max_{\{\Phi_t\}_{t=1}^H}\quad \E_{z_1,z_2,\cdots, z_H}[\sum_{t=1}^H r^\pi(s_t;\Phi_t(\theta))],\tag{OMAL}\label{eq:omal}\\
    &\text{s.t. }\nonumber\\
    &\qquad \qquad z_{t+1}\sim p_z(\cdot|z_t),\ t=1,\ldots, H-1,\nonumber\\
    &\qquad \qquad \theta=\argmax \E_{z\sim \rho_z}[J_z(\theta)]\nonumber.
\end{align}

In this context, the mode denoted as $z\in \mathcal{Z}$ represents the specific environment in which the offline policy is situated. Furthermore, we denote the latent mode transitions probabilistically via a Markov chain $p_z(z_{t+1}|z_t)$ with an initial distribution denoted as $\rho_z(z_1)$. Our proposition involves the adoption of the Conjectural Online Lookahead Adaptation (COLA) model, as outlined in \cite{cola} and expounded upon in \Cref{sec:method_cola}, as a means to identify a viable $\Phi_t$. The diagram illustration can also be found in \Cref{fig:system_diagram}.

\begin{figure}[h]
    \centering
    \includegraphics[width=3in]{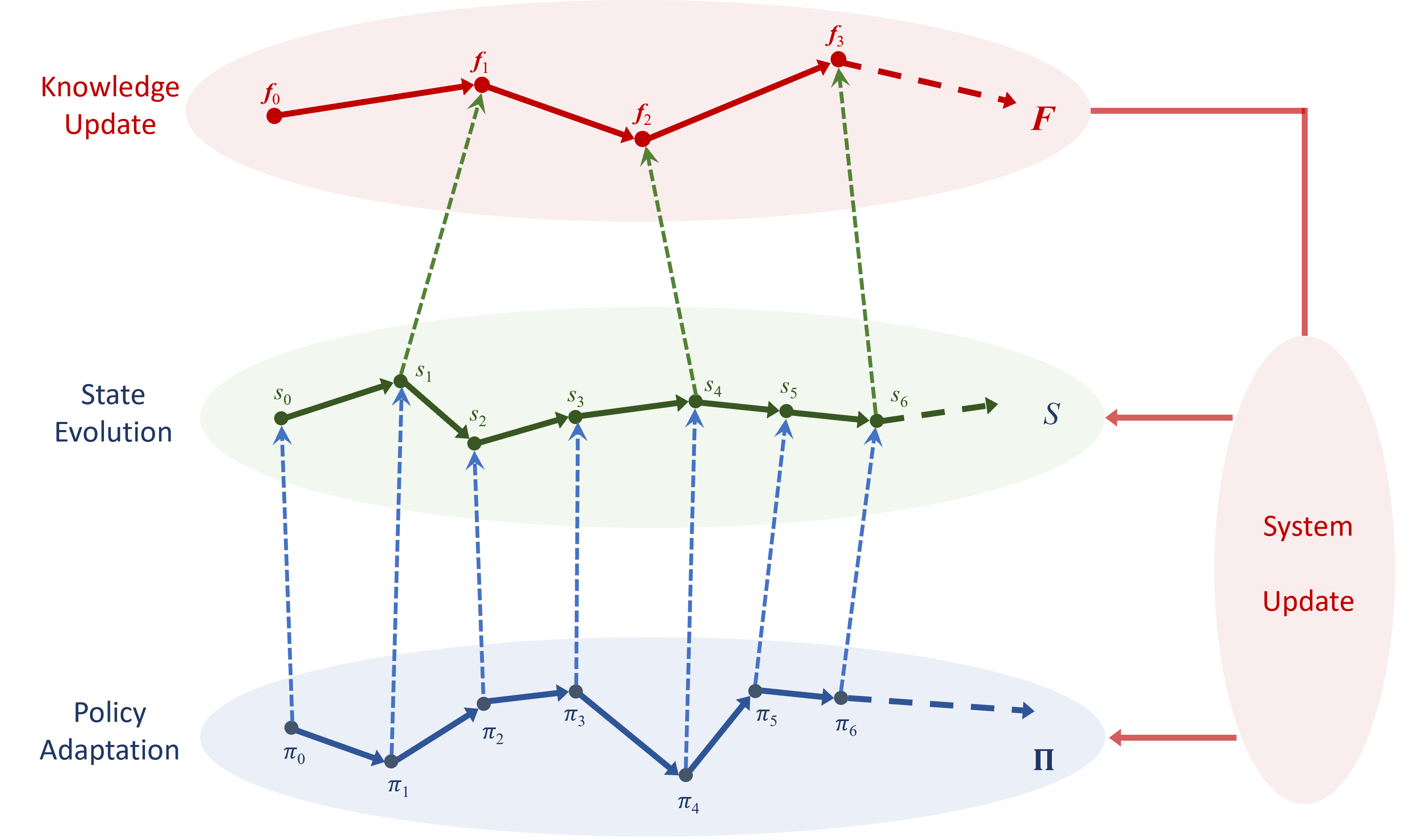}
    \caption{The NUMERLA framework is shown in the plot. Utilizing symbolic logic-based safety constraints and online meta-adaptation learning techniques, we consider the following scenario: With the state denoted as $s_t$, policy as $\pi_t$, and policy constraint as $f_i$ at time $t$, generated by the SSC function. The policy $\pi_t$ initiates dynamic adjustments online while guided by the knowledge encoded in $f_i$. This adaptation process draws upon insights from both the current state $s_t$ and historical context. Subsequently, the revised policy $s_{t+1}$ governs the selection of an action, thereby leading to the transition from state $s_t$ to $s_{t+1}$. Assuming the environment mode space $\mathcal{Z}_t$ changes exclusively during steps $1,4,6$. In case of such mode changes, the knowledge content is updated to $f_{i+1}$.}
    \label{fig:NUMERLA}
\end{figure}

\subsection{Objective function of OMAL with an action constraint}\label{sec:obj_NUMERLA}

In the last section, we address the OMAL problem that can be solved by COLA, which can be thought of as Neuro Lookahead Learning. This section explains how symbolism can make the policy safer. When the OMAL wants to find an optimal model that maximizes the online performance, the aim of NUMERLA is to further make sure $K$ steps safety of the policy. The objective function is given by:
\label{eq:safe-rl}
  \begin{align}
    &\max_{\{\Phi_t\}_{t=1}^H}\quad \E_{z_1,z_2,\cdots, z_H}[\sum_{t=1}^H r^\pi(s_t;\Phi_t(\theta))],\tag{NUMERLA}\label{eq:NUMERLA}\\
    &\text{s.t. }\nonumber\\
    &\qquad \qquad z_{t+1}\sim p_z(\cdot|z_t),\ t=1,\ldots, H-1,\nonumber\\
    &\qquad \qquad \theta=\argmax \E_{z\sim \rho_z}[J_z(\theta)]\nonumber,\\
    &\qquad \qquad \Phi_t(\theta) \in f_t(z_t),\nonumber\\
    &\qquad \qquad f_t(z_t):= \begin{cases}\varphi_1 & \text { if } \chi_1(z_t) \\ \varphi_2 & \text { if } \chi_2(z_t) \wedge \neg \chi_1(z_t) \\ \cdots & \\ \varphi_n & \text { if } \chi_n(z_t) \wedge\left(\bigwedge_{1 \leq i<n} \neg \chi_i(z_t)\right)\end{cases}.\tag{SSC}\nonumber
  \end{align}
where symbolic safety constraints (SSC) $f_t: \mathcal{Z}_t \Rightarrow \Theta_t$ is a function belonging to the space $\mathcal{F}$. The function $f_t$ serves to associate a mode within $\mathcal{Z}_t$ with a subset of $\Theta_t$. It is important to note that when alterations occur in the mode space $\mathcal{Z}_t$ at step $t$, corresponding adjustments are made to both the mapping function $f_t$ and the policy space $\Theta_t$ in accordance with the change. We define $\mathcal{X}:= \{\chi_1, \ldots, \chi_n\}$ as a collection of symbolic logic judgments (expressed through linear predicates), which serve to segment the space of modes. For the sake of clarity, we represent the non-overlapping partitions as $\{g_1, \cdots, g_n\}$, denoted by for all $i\in\{1,\cdots,n\}$, $z'\in g_i\subseteq \mathbf{Z}_t$ is a set of mode that satisfies $\mbox{if }\chi_i(z') \wedge \left(\bigwedge_{1 \leq j < i} \neg \chi_j(z')\right)\mbox{is true}$; $\{\varphi_1,\cdots, \varphi_n\}\subseteq \Theta$ are the symbolic logic-based safety constraints which are the coupling between the knowledge mode space with the physical action space. They can be defined as subsets in $\Theta$ that include the safest action choices according to the yield environment mode $z_t$. The framework of NUMERLA is also shown in \Cref{fig:NUMERLA}.


\section{Methodology of Optimization}\label{sec:method}
\subsection{Conjectural Online Lookahead Adaptation}\label{sec:method_cola}
Following the model in \cite{cola}, let $b_t$ be the agent's belief (normally, the belief is a pre-defined prediction or conjecture of the future mode in the environment) and $\theta$ still be our obtained policy defined in \Cref{eq:omal}. We consider a $K$ step future that can be represented by trajectory $\tau_t^{K}:=(s_t,a_t,\ldots,s_{t+K-1}, a_{t+K-1}, s_{t+K})$. Following, the distribution of trajectory $\tau_t^{K}$ can be characterized as:
\begin{align*}
&q(\tau_t^K;b_t,\theta):=\\
&\prod_{k=0}^{K-1}\pi(a_{t+k}|s_{t+k};\theta) \prod_{k=0}^{K-1}\left[\sum_{z\in \mathcal{Z}}b_t(z)\underbrace{P(s_{t+k+1}|s_{t+k},a_{t+k};z)}_{\text{unkown}}\right].
\end{align*}
Here, the transition of the environment $P$ is unknown.

The goal of this model is to maximize the forecast future performance:
\begin{align}
    \max_{\theta'\in \Theta}\E_{q(\tau_t^K;b,\theta')}\sum_{k=0}^{K-1} r(s_{t+k},a_{t+k})\label{eq:COLA_opt}
\end{align}
However, the agent cannot access the distribution $q(\tau_t^K;b,\theta')$ during the online adaptation. Thus, cannot use policy gradient methods to solve the optimization problem.

As the replacement, we use importance sampling to do the optimization by reformulating the original problem \eqref{eq:COLA_opt} to the conjectural lookahead optimization (CLO) problem:
\begin{align}\label{eq:trpo}
		\max_{\theta'\in\Theta} & \ \E_{q(\cdot;b_t,\theta)}\left[\prod_{k=0}^{K-1}\frac{\pi(a_{t+k}|s_{t+k};\theta')}{\pi(a_{t+k}|s_{t+k};\theta)}\sum_{k=0}^{K-1} r(s_{t+k},a_{t+k})\right]\tag{CLO}\\
	&\text{s.t. } \quad \E_{s\sim q} D_{KL}(\pi(\cdot|s;\theta), \pi(\cdot|s;\theta'))\leq \delta,\nonumber
\end{align}
where $D_{KL}$ is the Kullback-Leibler divergence. In the KL divergence constraint, we slightly abuse the notation $q(\cdot)$ to denote the discounted state visiting frequency $s\sim q$.

\Cref{eq:trpo} is equivalent to \Cref{eq:COLA_opt} since the distribution difference between $q(\tau_t^K;b,\theta')$ and $q(\tau_t^K;b,\theta)$ in \eqref{eq:trpo} is compensated by the ratio $\prod_{k=0}^{K-1}\frac{\pi(a_{t+k}|s_{t+k};\theta')}{\pi(a_{t+k}|s_{t+k};\theta)}$. When $\theta'$ is close to the based policy $\theta$ in terms of KL divergence, we can use the data collected during the training to finish the approximation of the results. In the COLA setting, the data is gradient sampling of the objective function in different environment modes. The overall online updating process for the COLA is shown in \Cref{algo:cola}.
\begin{algorithm}[H]	
	\caption{\textbf{C}onjectural \textbf{O}nline \textbf{L}ookahead \textbf{A}daptation}\label{algo:cola}
		\begin{algorithmic}
			\State  \textbf{Input} The meta policy $\theta$, belief $b$, training samples $\{\mathcal{D}_z\}$, sample batch size $M$, lookahead horizon $K$
                \For {$t\in \{1,2,\ldots, \}$}
			\State \textbf{Acquire} the sensor input $s_t$;
			\State \textbf{Implement} the action using $\pi(\cdot|s_t;\theta_t)$;
			\State \textbf{Update} the belief $b(z;s_t)$; 
			\State \textbf{Sample} $M$ trajectories ($K$ steps from $t$ ) $\hat{\tau}_t^K$ under  ${z}$ from $\{\mathcal{D}_z\}$;
			\State \textbf{Obtain} $\theta'$ by solving Conjecture Lookahead Optimization \eqref{eq:trpo}; 
                \State $\theta_{t+1}=\theta'$.
			\EndFor
		\end{algorithmic}
\end{algorithm}

\subsection{Symbolic Safety Constraint Adaptation}

\begin{figure*}[h]
    \centering
    \includegraphics[width=7.0in]{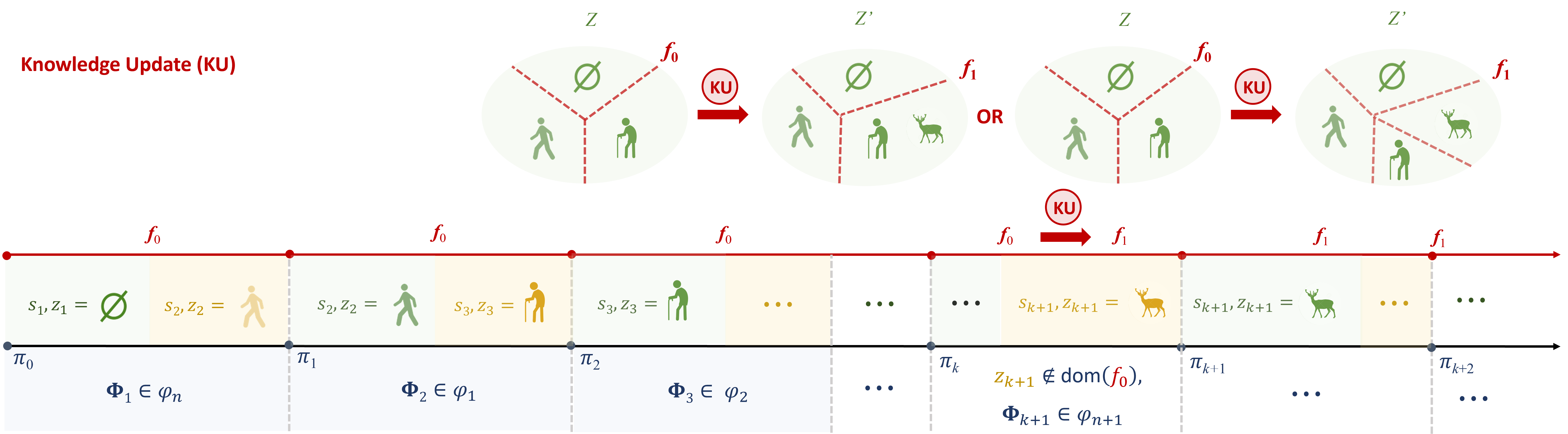}
    \caption{The evolution of the SSC function takes place through the absorption of new information. Suppose the initial SSC function is $f_0$. We assume for time step $1$ to $k$, $f_0$ can dominate everything. At $t=1$, the SSC uses $\varphi_n$ as its constraint since $z_1\in g_n$. The lookahead procedure conjectures the next time step should be in mode $z_2\in g_1$, so the SSC will prepare to use $\varphi_1$ as the next constraint. The knowledge update of SSC occurs when a novel mode is identified at $t=k$, denoted as $z_{k+1}\notin g_i$ for all existing modes $g_i$ within the set $\{g_1, \ldots, g_n\}$, or in other words, $z_{k+1}\notin \mbox{dom}(f_0)$. This update can be executed through two distinct approaches: either by integrating the new mode with an existing earlier mode (solving \Cref{eq:safe_p} with $g_i, \mbox{ }\forall i\in \{1,\cdots,n\}$) or by establishing a fresh partition exclusively for the new mode (solving \Cref{eq:safe_p} with $g_{n+1}$).}
    \label{fig:S_time}
\end{figure*}

Denote a given safety assessment function $Safe(s_t,a_t): S\times \mathcal{A}\rightarrow \{0,1\}$ that outputs a Boolean value where if state-action pair (s-a pair) $(s_t, a_t)$ is safe (output 0) or unsafe (output 1). Then, for the symbolic safety constraint adaptation (SSCA), its objective function can be defined as \Cref{eq:safe} shown below:

\begin{align}
    \min_{f} \sum_{z\in\mathcal{Z}}\sum_{\theta'\in f_t(b_t(z))}\mathbb{E}_{q(\tau_t^K;b,\theta')} \left[ \sum_{k=0}^{K-1} Safe(s_{t+k},a_{t+k}) \right].\tag{SSCA}
    \label{eq:safe}
\end{align}

On the other hand, we can divide it into optimization problems according to different mode partitions $g_i$, called the symbolic safety constraint adaptation of partition (SSCAP). Suppose we have:
\begin{align*}
&\hat{q}(\tau_t^K;b_t,g_i,\theta):=\\
&\prod_{k=0}^{K-1}\pi(a_{t+k}|s_{t+k};\theta) \prod_{k=0}^{K-1}\left[\sum_{z\in g_i}b_t(z)P(s_{t+k+1}|s_{t+k},a_{t+k};z)\right].
\end{align*}
Then, we can denote the SSC optimization for specific partition:
\begin{align}
    \min_{\varphi_i} \sum_{\theta'\in \varphi_i}\mathbb{E}_{\hat{q}(\tau_t^K;b_t,g_i,\theta)} \left[ \sum_{k=0}^{K-1} Safe(s_{t+k},a_{t+k}) \right].\tag{SSCAP}
    \label{eq:safe_p}
\end{align}

The foundational SSC function $f_0$ is derived through a heuristic process rooted in prior human insights within our conceptual framework. It is important to note that $\mathcal{Z}$ represents the range of modes entirely encompassed by $f_0$. However, in scenarios where the agent is confronted with a novel mode space denoted as $\mathcal{Z}'\supset\mathcal{Z}$ demanding a more powerful SSC function, a knowledge expansion is imperative. This expansion pertains to the enhancement of our understanding, specifically the SSC function, to effectively accommodate this broader mode space. The online update of the SCC can follow the rules described in \Cref{algo:SSC}.
\begin{algorithm}
  \caption{\textbf{S}ymbolic \textbf{S}afety \textbf{C}onstrain \textbf{A}daptation}\label{algo:SSC}
  \begin{algorithmic}[1]
    \State\textbf{Input} $\{\chi_1, \ldots, \chi_n\},\{\varphi_1,\cdots, \varphi_n\}, \mathcal{Z}'$
      \State \textbf{Create} partitions $\{g_1, \ldots, g_n\}$ using $\{\chi_1, \ldots, \chi_n\}$
      \State $g_{n+1} \gets \emptyset$
      \For{$z' \in \mathcal{Z}'$}
        \If {$z'\notin g_i, \forall g_i\in \{g_1, \ldots, g_n\}$}
        \State $g_{n+1} \gets g_{n+1} \cup \{z'\}$
        \EndIf
      \EndFor
      \State \textbf{Find} $g_{n+1}=\chi_{n+1}(z_t) \wedge\left(\bigwedge_{1 \leq i<n+1} \neg \chi_i(z_t)\right)$
      \State \textbf{Obtain} $\varphi_{n+1}$ that optimal \eqref{eq:safe_p} with input $p_{n+1}$
      \State \textbf{Derive} updated judgments $\{\chi_1, \ldots, \chi_{n+1}\}$
      \State \Return $\{\chi_1, \ldots, \chi_n, \chi_{n+1}\}$ and $\{\varphi_1,\cdots, \varphi_n,\varphi_{n+1}\}$
  \end{algorithmic}
\end{algorithm}

\Cref{fig:S_time} shows the process of online updating of the SSC function. By combining the results derived from \Cref{algo:SSC}, we acquire the refined SSC function denoted as $f_1$. It is essential to note that the enhancement of the SSC function is not a solitary, instantaneous modification; rather, the agent is required to gather data from the changing environment $\mathcal{Z}'$. This necessitates conducting multiple samplings from the environment to achieve the desired refinement. 

Illustrative examples can shed light on the process of updating the SCC. A relevant exasmple is motivated by the disparities in driving practices across different regions within the United States. Imagine a driver who has been accustomed to the driving conditions in New York City but relocates to Texas. This relocation exposes the driver to a distinct environmental context. In urban traffic settings, the driver's existing knowledge might still prove effective. However, driving in Texas introduces new scenarios, such as encountering wildlife like deer or bears on the road. Here, the driver not only adapts through personal experience but can also seek insights from local residents, or the acquisition of new modes online. Regarding the incorporation of these novel modes into the driver's cognitive framework, namely the expansion of SCC, this can be accomplished by making minor adjustments to an existing safety partition or creating a new partition catering exclusively to these new modes. These concepts are illustrated in Figure \ref{fig:S_time}.

In \Cref{sec:exp}, our focus is only on the scenario where the SCC function remains invariant throughout.

\section{Experimental Configuration}\label{sec:exp}
For our experimental assessments, we employ CARLA-0.9.4 \cite{carla}, a well-established platform for urban self-driving scenarios. To establish the communication between learning algorithms and environments, we adapt the API by integrating the Multi-Agent Connected Autonomous Driving (MACAD) Gym \cite{macad-gym} framework atop CARLA.

\begin{figure}[h]
    \centering
   \includegraphics[width=0.49\textwidth]{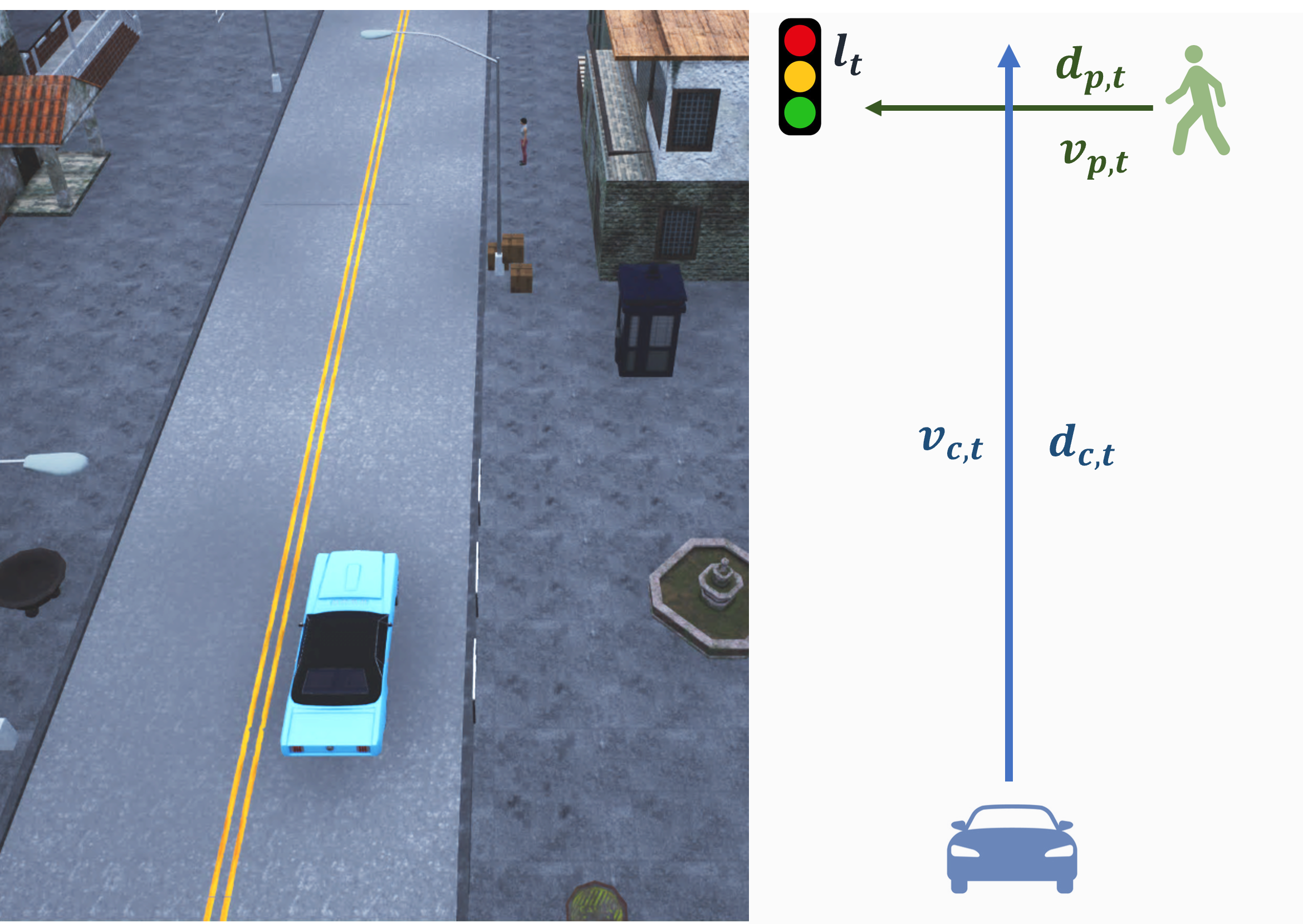}
    \caption{An illustration of the uncertain position of signal light scenario. In this, we create a pedestrian with a signal light in front of the car on the urban sidewalk road. The location of this pedestrian is uncertain. The sensors will observe the velocities $v_{c,t}, v_{p,t}$ and their distances to the destination $d_{c,t},d_{p,t}$ of the pedestrian and the vehicle and the signal light's status $l_t$. The vehicle needs to reach its destination in a short period of time without colliding with pedestrians.}·
    \label{fig:scenario_true}
\end{figure}

We examine vehicle-human interactions in an urban traffic environment featuring two agents: a vehicle with an initial velocity and a pedestrian, illustrated in \Cref{fig:scenario_true}. We denote the vehicle by $c$ and the pedestrian $p$. To assess the effectiveness of our approach, we conduct experiments within two distinct scenarios: one involving Well-Behaved walking and the other involving jaywalking. Each scenario comprises three tasks determined by the initial distance between the vehicle's and pedestrian's origin points. We will describe more specific details later.

We assume the state input is coming from the sensors on the vehicle. The state representation comprises each agent's current and previous speeds, denoted as $v_{c,t}, v_{p,t} \in \mathbb{R}$, and their distances to their respective endpoints, represented by $d_{c,t}, d_{p,t} \in \mathbb{R}$. Additionally, the actions $a_{c,t}\in \mathcal{A}_c\subseteq \mathbb{R}^{n}$ and $a_{p,t}\in \mathcal{A}_p\subseteq \mathbb{R}^{n}$ are included. Furthermore, we introduce a simulated signal light input denoted as $l_t$, which serves as an additional component within the state. It is important to highlight that, since the sensors are only equipped on the vehicle, inputs stemming from pedestrians and the signal light are initialized to $-1$ until the vehicle approaches within a distance of $15$ meters from them. The complete structure of the state $s_t$ encompasses 10 different variables, namely $\{d_{c,t}, d_{p,t}, v_{c,t}, v_{p,t}, l_t, d_{c,t-1}, d_{p,t-1}, v_{c,t-1}, v_{p,t-1}, l_{t-1}\}$. When executing the SSC function $f_t$, we focus solely on the current state information, represented as $\hat{s}_t=\{d_{ct}, d_{pt}, v_{ct}, v_{pt}, l_t\}$, to ensure computational efficiency. 

For pedestrians and the vehicle agent, the available actions are defined in \Cref{tb:actions_set}. In the case of pedestrians, the action values correspond to acceleration towards the main road direction (if positive) or the opposite direction (if negative). For the vehicle, the action values represent the throttle strength (if positive) or the brake strength (if negative).

\begin{table}[h]
\centering
\caption{Discrete Actions}
\begin{tabular}{|c|c|c|c|}
\hline
\textbf{\#} & \textbf{Action} & \textbf{\#} & \textbf{Action} \\
\hline\label{tb:actions_set}
0 & 0.0 & 4 & -1.0 \\
\hline
1 & 1.0 & 5 & -0.5 \\
\hline
2 & 0.5 & 6 & -0.25 \\
\hline
3 & 0.25  & & \\
\hline
\end{tabular}
\end{table}

The vehicle's reward function hinges on its present velocity, proximity to the destination, and the occurrence of collisions. Across every scenario, encompassing both Well-Behaved walking and jaywalking scenarios, we sketch three distinct initial gaps (15 meters, 25 meters, and 35 meters) between the vehicle and pedestrians, classified according to their types. This methodology serves to assess how effectively and safely the proposed NUMERLA model can adeptly manage diverse traffic scenarios, thereby measuring its adaptability and security.

In both scenarios, we assess the performance of the RL method, COLA method, and NUMERLA method for each individual task. We capture the mean reward, standard deviation, and collision rate as key metrics for each experimental iteration.

\subsection{Well-Behaved Walking}

\begin{figure}[h]
    \centering
    \includegraphics[width=3in]{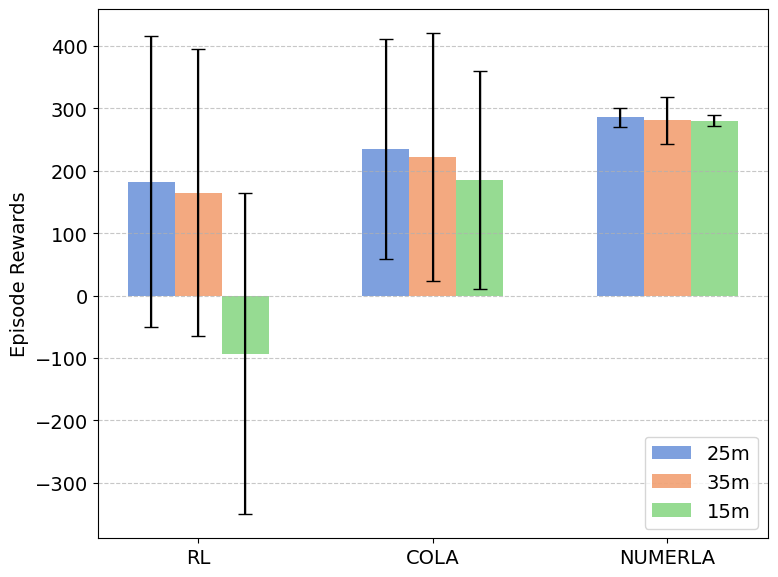}
    \caption{The performance comparison between RL, COLA, and NUMERLA for Well-Behaved walking pedestrians where the value represents the mean rewards and the error bar represents the standard deviation (std). The data is gathered from 1,000 episodes of online executions. The RL performance is the worst of the three types of methods, while the COLA obtains some better results. However, both of these two methods return us to a poor std, which means unstable performance. By using the NUMERLA method, we can achieve higher mean rewards and a small std. It should be noted that the task 15-meter gets the worst performance in every method. The reason is that the 15-meter is the hardest task in this urban environment since our vehicle has an initial speed, but the location of the pedestrians is too close.}
    \label{fig:performance}
\end{figure}

In this scenario, the behavior of the pedestrian is guided by the signal light. When the signal light is red, the pedestrian refrains from initiating movement. When the signal light turns yellow, there is a 0.1 probability that the pedestrian will commence walking. When the signal light switches to green, the pedestrian promptly begins walking.

The \Cref{fig:performance} shows the efficiency and long-term safety performance of the NUMERLA method compared with the RL and the COLA in the Well-Behaved walking scenario. We collect the collision rate, which means the ratio of episodes with collision and the testing episode number, shown in \Cref{tab:collision-rates}. We can find collision rates are around zero for the NUMERLA method, which is much safer than the other two methods.

\begin{table}[ht]
    \centering
    \begin{tabular}{|c|c|c|c|}
        \hline
        \multirow{2}{*}{Policy Type} & \multicolumn{3}{c|}{Collision Rate} \\ \cline{2-4}
        & 25m & 35m & 15m \\ \hline
        RL & 0.251 & 0.303 & 0.718 \\ \hline
        COLA & 0.091 & 0.113 & 0.201 \\ \hline
        NUMERLA & 0.000 & 0.0004 & 0.000 \\ \hline
    \end{tabular}
    \caption{Collision Rates for Well-Behaved Walking}
    \label{tab:collision-rates}
\end{table}

\subsection{Jaywalking}

\begin{figure}[h]
    \centering
    \includegraphics[width=3in]{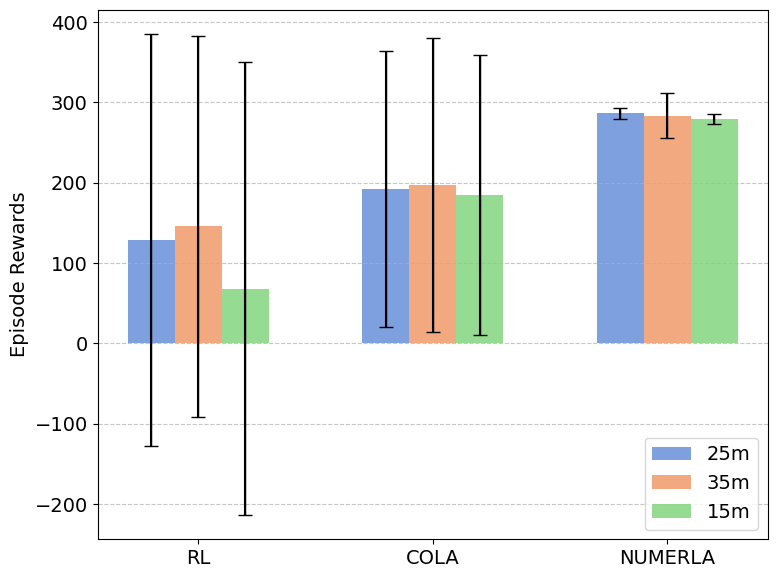}
    \caption{The performance comparison between RL, COLA, and NUMERLA for jaywalking pedestrians where the value represents the mean rewards and the error bar represents the std. The data is gathered from 1,000 episodes of online executions. We should notice that all std get even bigger than the experiments in Well-Behaved Walking since the pedestrians are unpredictable under this circumstance. However, the performance of the reinforcement learning (RL) technique excels in the 15-meter task compared to the compliant walking scenario. This is attributed to pedestrians having the flexibility to decide whether or not to cross the street, even opting for a significantly delayed crossing if desired. But, our NUMERLA method still can maintain a high mean reward and small std performance in this scenario.}
    \label{fig:performance_random}
\end{figure}

In this scenario, the behavior of the pedestrian is unpredictable. The pedestrian chooses a random time to initiate walking, irrespective of the ongoing color of the signal light.

The \Cref{fig:performance_random} shows the efficiency and long-term safety performance of the NUMERLA method compared with the RL and the COLA, even though the pedestrians' behavior is unpredictable in this scenario. In \Cref{tab:collision-rates_random}, we can also find collision rates are still around zero for the NUMERLA method, showing the better performance of our method.
\begin{table}[ht]
    \centering
    \begin{tabular}{|c|c|c|c|}
        \hline
        \multirow{2}{*}{Policy Type} & \multicolumn{3}{c|}{Collision Rate} \\ \cline{2-4}
        & 25m & 35m & 15m \\ \hline
        RL & 0.350 & 0.341 & 0.438 \\ \hline
        COLA & 0.156 & 0.154 & 0.190 \\ \hline
        NUMERLA & 0.000 & 0.0003 & 0.000 \\ \hline
    \end{tabular}
    \caption{Collision Rates for Jaywalking}
    \label{tab:collision-rates_random}
\end{table}


\section{Conclusion}
This work has introduced a novel online meta-learning approach, building upon the principles of Neurosymbolic Meta-Reinforcement Lookahead Learning (NUMERLA). This technique guarantees the security of real-time learning by consistently refining safety constraints. NUMERLA enables long-term safe online adaptation by solving the conjectural lookahead optimization (CLO) and the symbolic safety constraint adaptation (SSCA) on the fly using off-policy data and the conjecture of the future.

\bibliographystyle{IEEEtran}
\bibliography{NUMERLA}



\end{document}